\documentclass{article} 
\usepackage{tccml_iclr2025_conference,times}

\usepackage{amsmath,amsfonts,bm}

\def\eqref#1{equation~\ref{#1}}

\def\1{\bm{1}}

\DeclareMathAlphabet{\mathsfit}{\encodingdefault}{\sfdefault}{m}{sl}
\SetMathAlphabet{\mathsfit}{bold}{\encodingdefault}{\sfdefault}{bx}{n}

\usepackage{hyperref}
\usepackage{url}
\usepackage{algorithm}
\usepackage{algpseudocode}
\usepackage{amsmath}
\usepackage{graphicx}
\usepackage{wrapfig}
\usepackage{svg}
\usepackage{enumitem}   

\title{Offline Reinforcement Learning for \\ Microgrid Voltage Regulation}

\author{Shan Yang \\
School of System Science and Engineering \\
Sun-Yat Sen University \\
Guangzhou, China \\
\texttt{yangsh237@mail2.sysu.edu.cn} \\
\And
Yongli Zhu\thanks{Corresponding author.} \\
School of System Science and Engineering \\
Sun-Yat Sen University \\
Guangzhou, China \\
\texttt{yzhu16@alum.utk.edu} \\
}

\algrenewcommand\algorithmicrequire{\textbf{Input:}}
\algrenewcommand\algorithmicensure{\textbf{Output:}}

\iclrfinalcopy 
\begin{document}

\maketitle

\begin{abstract}
This paper presents a study on using different offline reinforcement learning algorithms for microgrid voltage regulation with solar power penetration. When environment interaction is unviable due to technical or safety reasons, the proposed approach can still obtain an applicable model through offline-style training on a previously collected dataset, lowering the negative impact of lacking online environment interactions. Experiment results on the IEEE 33-bus system demonstrate the feasibility and effectiveness of the proposed approach on different offline datasets, including the one with merely low-quality experience.

\end{abstract}

\section{Introduction}
\label{sec:intro}

The rapid integration of distributed energy resources (DERs), such as photovoltaic (PV) systems, has introduced significant challenges in maintaining stable voltage levels within power distribution networks~\citep{tonkoski2012impact}. Unlike traditional centralized power systems, modern distribution networks, e.g., microgrids with multiple distributed renewable sources, are more dynamic and decentralized~\cite {ali2024dynamic, fusco2021decentralized}, making voltage regulation particularly challenging. Advanced control strategies are required to address fluctuations caused by changes in load demand and intermittent renewable generation~\citep{wei2024online, singhal2018real}.

Reinforcement learning (RL) has emerged as a powerful tool for optimizing control policies in complex and dynamic environments~\citep{cao2020reinforcement}. Its ability to learn from past experiences and adapt to diverse scenarios makes it an appealing solution for voltage regulation. However, standard RL methods, such as Q-learning and Deep Deterministic Policy Gradient (DDPG), typically rely on real-time environmental interactions to explore and improve their policies. While effective, this reliance on real-time interactions makes these methods challenging to apply in real-world power systems, where online experimentation can be costly, risky, and potentially destabilizing~\citep{levine2020offline, schmidt2024learning}.

Offline reinforcement learning (Offline RL) trains agents on pre-collected datasets, reducing the need for real-time interactions~\citep{chen2024opportunities, hu2024multi}. However, it faces challenges like extrapolation errors when evaluating out-of-distribution actions, leading to overly optimistic Q-values and unstable policies~\citep{levine2020offline}. These risks are critical in safety-sensitive tasks like voltage regulation~\citep{zhang2024application}. To address this, algorithms such as batch-constrained deep Q-learning (BCQ) constrain actions to remain close to the offline dataset, reducing the likelihood of high-risk decisions~\citep{fujimoto2019off}.

This study compares the BCQ and CQL~\citep{kumar2020conservative} algorithms in addressing the voltage regulation problem in a PV-penetrated distribution network. Section~\ref{sec:offline_rl} details the architectures of both the BCQ and CQL algorithmic frameworks, highlighting their respective mechanisms for mitigating extrapolation errors. Section~\ref{sec:dn-vr} describes the offline dataset creation and collection used in this study, including datasets of varying quality levels. Section~\ref{sec:experiments} outlines the experimental setup and presents simulation results on the IEEE 33-bus system, focusing on the performance of BCQ and CQL across different datasets. Finally, Section~\ref{sec:conclusion} provides conclusions and discusses potential future work in this research area.

\section{Offline Reinforcement Learning}
\label{sec:offline_rl}

\subsection{Basic Idea}
In scenarios such as voltage regulation of a microgrid, direct experimentation with the live system is often impractical due to cost, safety, and legal concerns.
\vspace{-5pt}
\begin{figure}[H]
    \centering
    \includegraphics[width=1\textwidth]{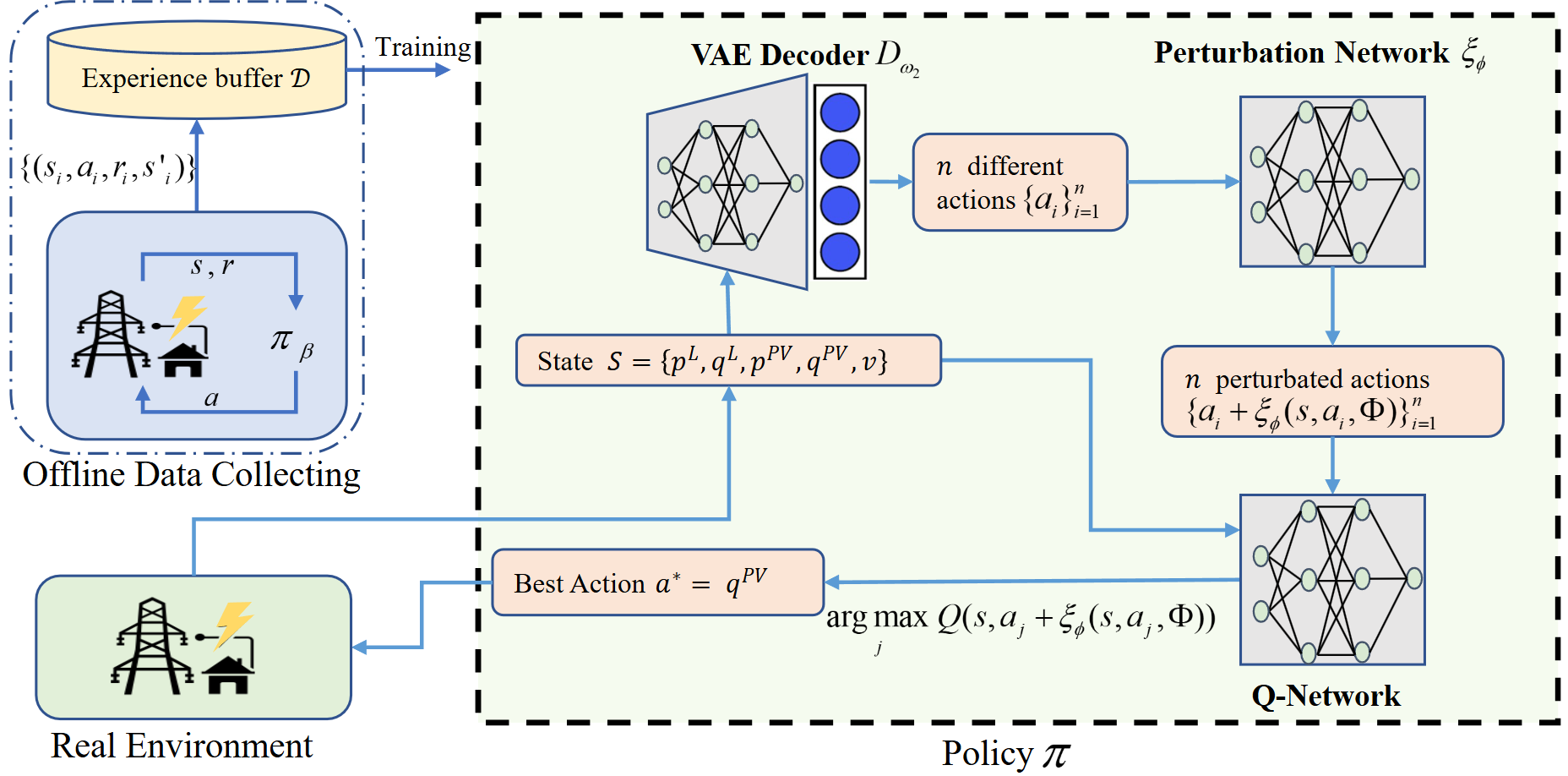}
    \vspace{-12pt}
    \caption{The basic idea of offline reinforcement learning. }
    \label{fig:offline_rl_idea}
\end{figure}
\vspace{-15pt}

Offline reinforcement learning addresses these challenges by training solely on pre-collected datasets
(\(\mathcal{D} = \{(s^i, a^i, r^i, s_{i+1}^i)\}\), where transitions are gathered by a behavior policy \(\pi_{\beta}\)). 
This data-driven approach is suitable for settings where system safety and stability are critical and typically involves three steps:

\begin{enumerate}[label=(\roman*)]
    \item Data collection using a fixed (or simple) behavior policy;
    \item Offline training with the collected dataset;
    \item Deployment of the trained policy in the real environment.
\end{enumerate}

This approach allows the agent to develop a policy that meets control goals while avoiding the risks of direct interaction with the system.
\subsection{The BCQ algorithm}
\label{subsec:bcq}

The BCQ algorithm intends to maximize the expected cumulative reward under a constraint that prevents the agent from selecting actions significantly different from those observed in the offline dataset. In an offline setting, this is crucial for preventing the agent from overestimating Q-values for unfamiliar state-action pairs. Given a dataset
\(\mathcal{D}\) collected by a behavior policy \(\pi_{\beta}\), the goal of BCQ is to learn a policy \(\pi\) that maximizes:
\begin{equation}
    \label{eq:bcq-objective}
    \mathbb{E}_{(s,a,r,s')\sim \mathcal{D}}\!\Bigl[r + \gamma \max_{a' \in \text{Batch}} Q(s', a')\Bigr]
\end{equation}
where \(\gamma\) is the discount factor, and ``Batch'' refers to the set of actions close to the distribution in \(\mathcal{D}\).

To further constrain the action selection and improve generalization, BCQ uses a state-conditioned generative model, namely a variational autoencoder (VAE) (denoted by \(G_{\omega}(s)\)), to approximate feasible actions for each state. The VAE is trained to generate actions that are likely under the behavior policy \(\pi_{\beta}\), filtering out actions that deviate too far from the observed data. This setup effectively restricts the agent’s exploration to actions close to those in the offline dataset, mitigating the risk of extrapolation errors. More algorithmic details can be found in Appendix \ref{app:bcq_algo}.

\subsection{The CQL Algorithm}
Conservative Q-Learning (CQL) is an offline reinforcement learning algorithm designed to mitigate extrapolation errors by modifying the Q-function's loss function. Extrapolation errors occur when the Q-function assigns overly optimistic values to actions that are poorly represented or absent in the dataset, leading to unreliable or unsafe policies. To address this, CQL introduces an additional penalty term to the Q-value loss, ensuring conservative estimates and prioritizing stable policy outcomes.

CQL optimizes the following objective by penalizing overestimation:
\begin{equation}
    J_{\text{CQL}}(Q) = \mathbb{E}_{s \sim \mathcal{D}} \Big[ \log \sum_{a} \exp Q(s, a) - \mathbb{E}_{a \sim \mathcal{D}}[Q(s, a)] \Big]
\end{equation}
where the first term penalizes high Q-values across the entire action space, preventing overestimation for unobserved actions. The second term prioritizes actions within the dataset distribution to ensure conservative policy behavior. More algorithmic details can be found in Appendix \ref{app:cql_algo}.

\section{Microgrid Voltage Regulation}
\label{sec:dn-vr}

\subsection{Environment}
The environment is a \texttt{distflow} program that simulates the power grid's power flow at each time step~\citep{wang2021multi}. It evaluates control algorithms under realistic operational conditions by simulating variations in load and PV generation. The environment includes state, action, and reward designs tailored for the voltage regulation task. For brevity, the detailed definitions of the state space, action space, reward function, and objective function are provided in Appendix~\ref{appendix:environment}.

\subsection{Data Collected by Different Levels of Policies}
We create three distinct datasets: \emph{Expert}, \emph{Medium}, and \emph{Poor}. Each dataset contains 200000 transitions $(s, a, r, s')$.

\vspace{-10pt}
\begin{figure}[ht]
    \centering
    \includegraphics[width=0.6\textwidth]{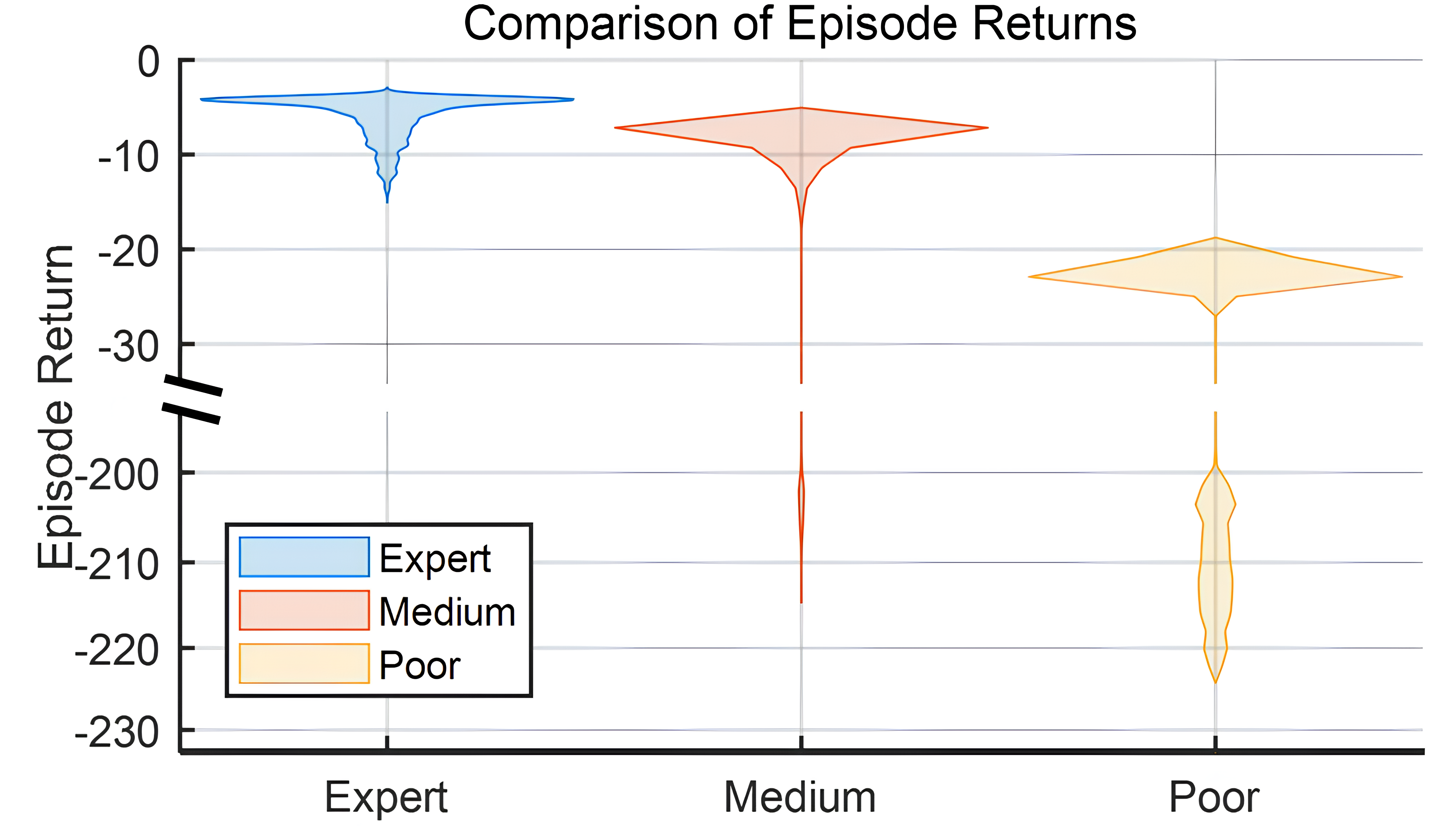}
    \vspace{-12pt}
    \caption{The episode returns across datasets.}
    
    \label{fig:episode_returns}
\end{figure}
\vspace{-10pt}

\textbf{Expert Dataset} 
This dataset is generated by a well-trained DDPG agent that consistently performs at a near-optimal level. The Expert dataset contains high-quality transitions with high average returns, simulating the scenario where the agent has a “good teacher”. This dataset benchmarks how well an agent can perform when provided with the best possible data.

\textbf{Medium Dataset}
Created by introducing Gaussian noise (5\%) to the Expert dataset’s actions, this dataset simulates a moderately effective policy. The added noise introduces variability, resulting in a mix of high and moderate-performance episodes. This dataset broadens the state-action pairs the agent encounters, testing its ability to learn from data that include suboptimal experiences.

\textbf{Poor Dataset}
This dataset \emph{deliberately} employs a poor policy by randomly selecting actions from the action space. It includes episodes of low average returns due to the lack of purposeful actions. This dataset pushes the offline-RL agent to extract useful patterns from poorly performed data as much as possible. It is used as a touchstone to inspect the genuine effectiveness of the offline RL algorithm before it is applied to the real system.

\section{Experiment Results}
\label{sec:experiments}

This section presents the experimental results of respectively applying the BCQ and CQL algorithms for the voltage regulation task on the IEEE 33-bus system (cf. Appendix~\ref{appendix:ieee33bus}). 

\begin{figure}[h] \centering \includegraphics[width=1\textwidth, trim={2.8cm 0cm 2.5cm 0cm}, clip]{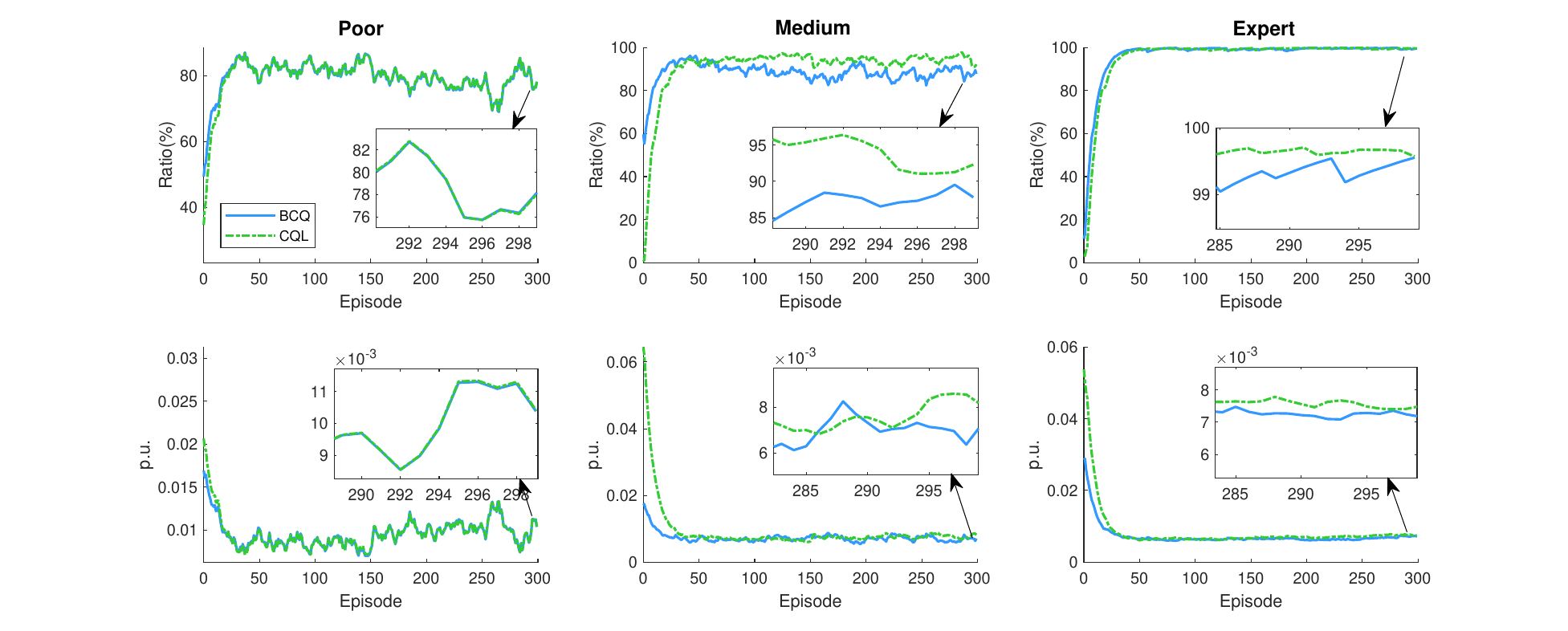} 
\vspace{-14pt}
\caption{Performance comparison of BCQ and CQL across three datasets: Totally Controllable Ratio (top row) and Average Voltage Deviation (bottom row).} \label{fig:bcq_vs_cql} \end{figure}

This section compares BCQ and CQL on three offline datasets: Poor, Medium, and Expert. The results, shown in Fig.~\ref{fig:bcq_vs_cql}, highlight each algorithm's performance in maintaining controllable voltage levels and minimizing voltage deviations. For more details on the performance of BCQ across the datasets, please refer to Appendix~\ref{appendix:bcq_performance}.

\vspace{-10pt}
\vspace{-5pt}

The comparison focuses on two key metrics: 1) \textit{Totally Controllable Ratio}: The percentage of time steps where all buses maintain voltages within the acceptable range (0.95 to 1.05 p.u.), reflecting controllability and stability. 2) \textit{Average Voltage Deviation}: The average deviation of bus voltages from the target level (1.0 p.u.), indicating voltage regulation effectiveness.

\textbf{Performance on Poor Dataset:}
CQL achieves higher controllable ratios and lower voltage deviations than BCQ, demonstrating robustness on low-quality data 

\textbf{Performance on Medium Dataset:}
CQL significantly outperforms BCQ, benefiting from its emphasis on the in-distribution actions, leading to better stability even with moderate data noise.

\textbf{Performance on Expert Dataset:}
Both algorithms perform well with high-quality data, achieving high controllable ratios and low deviations. CQL behaves slightly more stable.

In summary, CQL outperforms BCQ across all three datasets, particularly in scenarios with low and moderate data quality, making it a safer and more robust choice for voltage regulation tasks.

\section{Conclusion} \label{sec:conclusion}

This study investigates the application of Offline Reinforcement Learning (Offline RL) for voltage regulation in the PV-penetrated microgrid, focusing on BCQ and CQL algorithms. The results demonstrate that Offline RL can effectively utilize pre-collected datasets to learn stable and reliable control policies without requiring real-time interactions.

CQL consistently outperforms BCQ across datasets of varying quality, particularly in scenarios with low or medium-quality data, due to its conservative Q-value estimation that enhances robustness and stability. While BCQ provides a solid baseline, its performance is more sensitive to data quality.

Future research could explore applying these methods to larger and more complex power grids, integrating additional Offline RL approaches, and improving scalability to address real-world operational challenges in power systems.

\bibliographystyle{plainnat} 
\bibliography{iclr2025_conference}

\newpage
\appendix

\section{Appendix}
\subsection{Algorithm Details of BCQ and CQL}

\subsubsection{The BCQ algorithm}
\label{app:bcq_algo}

In the BCQ framework, the VAE outputs a set of candidate actions \(\{\,a_i \sim G_{\omega}(s')\}_{i=1}^{n}\) for each state \(s'\). These actions are then slightly perturbed to allow exploration within a safe neighborhood of the observed actions:
\begin{equation}
    \label{eq:perturbation}
    a_i \;=\; a_i + \xi_{\phi}\bigl(s', a_i, \Phi\bigr),
\end{equation}
where \(\xi_{\phi}\) is a perturbation model with a bound value \(\Phi\).

Finally, the Q-function in BCQ is updated by taking a conservative approach to avoid overestimation. The target value for a transition \((s, a, r, s')\) is calculated as:
\begin{equation}
    \label{eq:target}
    y \;=\; r \;+\; \gamma \max_{a_i\in\text{Batch}}
    \Bigl[\, \lambda \min_{j=1,2} Q_{\theta_{j}}(s', a_i)
    \;+\; (1-\lambda)\max_{j=1,2} Q_{\theta_{j}}(s', a_i)
    \Bigr],
\end{equation}
where \(\lambda\) controls the balance between pessimistic and optimistic value estimates, and \(Q_{\theta_j}\) denotes the target Q-network.

By integrating the VAE to generate action candidates close to the dataset’s behavior, BCQ ensures a conservative learning approach, enhancing its suitability for the offline voltage regulation task in power grid. The complete pseudocode for the BCQ algorithm is shown in Algorithm~\ref{alg:bcq}.

\begin{algorithm}[H]
\caption{Batch-Constrained Q-learning (BCQ)}

\label{alg:bcq}
\begin{algorithmic}[1]
\Require Batch $B$, horizon $T$, target update rate $\tau$, mini-batch size $N$, max perturbation $\Phi$, sampled actions $n$, weighting $\lambda$
\Ensure Trained networks for BCQ
\State Initialize $Q_{\theta_1}, Q_{\theta_2}, \xi_{\phi}, G_{\omega} = \{E_{\omega_1}, D_{\omega_2}\}$ with random parameters $\theta_1, \theta_2, \phi, \omega$
\State Initialize target networks $Q_{\theta'_1}, Q_{\theta'_2}, \xi'_{\phi}$: $\theta'_1 \gets \theta_1, \theta'_2 \gets \theta_2, \phi' \gets \phi$
\For{$t = 1$ to $T$}
    \State Sample $N$ transitions $(s,a,r,s')$ from $B$ and compute $\mu, \sigma = E_{\omega_1}(s,a), a' = D_{\omega_2}(s, z), z \sim \mathcal{N}(\mu,\sigma)$. Update $\omega \gets \arg\min_{\omega} \sum (a - a')^2 + D_{\mathrm{KL}}(\mathcal{N}(\mu,\sigma)\|\mathcal{N}(0,1))$.
    \State Sample $n$ actions $\{a_i \sim G_{\omega}(s')\}_{i=1}^n$, perturb $a_i$: $a_i \gets a_i + \xi_{\phi}(s', a_i, \Phi)$.
    \State Compute target $y = r + \gamma \max_{a_i} \Big[ (1 - \lambda) \max_{j=1,2} Q_{\theta'_j}(s', a_i) + \lambda \min_{j=1,2} Q_{\theta'_j}(s', a_i) \Big]$.
    \State Update $\theta_i \gets \arg\min_{\theta_i} \sum \big(y - Q_{\theta_i}(s, a)\big)^2$.
    \State Update $\phi \gets \arg\max_{\phi} \sum Q_{\theta_i}\big(s, a + \xi_{\phi}(s,a,\Phi)\big), a \sim G_{\omega}(s)$.
    \State Update target networks: $\theta'_i \gets \tau\,\theta_i + (1 - \tau)\,\theta'_i$, $\phi' \gets \tau\,\phi + (1 - \tau)\,\phi'$.
\EndFor
\end{algorithmic}
\end{algorithm}
\vspace{-10pt} 

\subsubsection{The CQL algorithm}
\label{app:cql_algo}

In the CQL framework, the following penalty is incorporated into the Q-function update (along with the standard Bellman loss) to make the training process more stable:
\begin{equation}
    L(Q) = J_{\text{CQL}}(Q) + \zeta \cdot \mathbb{E}_{(s, a, r, s') \sim \mathcal{D}} \Big[ \big( r + \gamma \max_{a'} Q(s', a') - Q(s, a) \big)^2 \Big]
\end{equation}
where \( \zeta \) balances the conservative penalty \( J_{\text{CQL}}(Q) \) and the Bellman consistency.

Furthermore, to encourage conservative behavior, CQL introduces an action-specific penalty term during policy evaluation, ensuring that the learned policy does not favor unobserved actions:
\begin{equation}
    \pi(a \mid s) \approx \exp \Big( Q(s, a) - \max_{a'} Q(s, a') \Big)
\end{equation}
where the softmax-like formulation adjusts action probabilities, giving higher weights to actions with lower penalties.

By combining these components, CQL effectively reduces the risk of unsafe policy execution, making it particularly suitable for offline reinforcement learning scenarios where live environment interaction is restricted. The complete pseudocode for the BCQ algorithm is shown in Algorithm~\ref{alg:cql}.

\begin{algorithm}[H]
\caption{Conservative Q-Learning (CQL)}
\label{alg:cql}
\begin{algorithmic}[1]
\Require Batch $B$, horizon $T$, target update rate $\tau$, mini-batch size $N$, max perturbation $\Phi$, sampled actions $n$, weighting $\lambda$ and $\zeta$
\Ensure Trained networks for CQL
\State Initialize $Q_{\theta_1}, Q_{\theta_2}, \xi_{\phi}, G_{\omega} = \{E_{\omega_1}, D_{\omega_2}\}$ with random parameters $\theta_1, \theta_2, \phi, \omega$
\State Initialize target networks $Q_{\theta'_1}, Q_{\theta'_2}, \xi'_{\phi}$: $\theta'_1 \gets \theta_1, \theta'_2 \gets \theta_2, \phi' \gets \phi$
\For{$t = 1$ to $T$}
    \State Sample $N$ transitions $(s,a,r,s')$ from $B$ and compute $\mu, \sigma = E_{\omega_1}(s,a), a' = D_{\omega_2}(s, z), z \sim \mathcal{N}(\mu,\sigma)$. Update $\omega \gets \arg\min_{\omega} \sum (a - a')^2 + D_{\mathrm{KL}}(\mathcal{N}(\mu,\sigma)\|\mathcal{N}(0,1))$.
    \State Sample $n$ actions $\{a_i \sim G_{\omega}(s')\}_{i=1}^n$, perturb $a_i$: $a_i \gets a_i + \xi_{\phi}(s', a_i, \Phi)$.
    \State Compute target $y = r + \gamma \max_{a_i} \Big[ (1 - \lambda) \max_{j=1,2} Q_{\theta'_j}(s', a_i) + \lambda \min_{j=1,2} Q_{\theta'_j}(s', a_i) \Big]$.
    \State Compute CQL penalty:
    \[
    J_{\text{CQL}}(Q_{\theta_i}) = \mathbb{E}_{s \sim B} \Big[ \log \sum_{a} \exp Q_{\theta_i}(s, a) - \mathbb{E}_{a \sim B}[Q_{\theta_i}(s, a)] \Big]
    \]
    \State Update \( Q \)-function with combined CQL penalty and Bellman loss:
    \[
    L(Q_{\theta_i}) = J_{\text{CQL}}(Q_{\theta_i}) + \zeta \cdot \mathbb{E}_{(s, a, r, s') \sim B} \Big[ \big( y - Q_{\theta_i}(s, a) \big)^2 \Big]
    \]
    \State Update $\theta_i \gets \arg\min_{\theta_i} L(Q_{\theta_i})$.
    \State Update $\phi \gets \arg\max_{\phi} \sum Q_{\theta_i}\big(s, a + \xi_{\phi}(s,a,\Phi)\big), a \sim G_{\omega}(s)$.
    \State Update target networks: $\theta'_i \gets \tau\,\theta_i + (1 - \tau)\,\theta'_i$, $\phi' \gets \tau\,\phi + (1 - \tau)\,\phi'$.
\EndFor
\end{algorithmic}
\end{algorithm}

\subsection{Environment Details}
\label{appendix:environment}

The environment is built on a \texttt{distflow} program that solves the microgrid's power flow at each time step~\citep{wang2021multi}. By simulating realistic variations in load demand and PV generation, it provides a rigorous framework to evaluate the performance of control algorithms.

\textbf{State Definition ($S$):} The state $S$ represents the current operational state of the microgrid and includes the following components:
\begin{itemize}
    \item Active and reactive power loads at each bus: $(p^L, q^L)$,
    \item Active and reactive power generated by PV inverters: $(p^{PV}, q^{PV})$,
    \item Voltage magnitudes at each bus: $\mathbf{v}$.
\end{itemize}
Mathematically, the state is defined as:
\begin{equation}
  \label{eq:state_appendix}
  S \;=\; \bigl\{\,p^L,\; q^L,\; p^{PV},\; q^{PV},\; \mathbf{v}\bigr\}
\end{equation}

\textbf{Action Definition ($A$):} The action $a_k$ corresponds to the control signal for the $k$-th PV inverter, representing the ratio of its maximum reactive power output:
\begin{equation}
  \label{eq:action_appendix}
  q_k^{PV} \;=\; a_k
    \cdot \sqrt{\bigl(S^{\max}_k\bigr)^{2} - \bigl(p_k^{PV}\bigr)^{2}}
\end{equation}
where $-c \leq a_k \leq c$, and $c$ is the maximum allowable ratio of reactive power injection or absorption. This design allows the agent to control voltage levels by adjusting the reactive power output of PV inverters.

\textbf{Reward Function ($r$):} The reward function incentivizes voltage stability and penalizes excessive reactive power usage:
\begin{equation}
    \label{eq:reward_appendix}
    r \;=\; -\frac{1}{\lvert V \rvert} \sum_{i \in V} \Bigl[\,
      \alpha_{1}\,\bigl\lvert v_{i}\bigr\rvert \;+\; \alpha_{2}\,\bigl\lvert q_{i}^{PV}\bigr\rvert
      \Bigr]
\end{equation}
where:
\begin{itemize}
    \item $\lvert v_{i}\rvert$ represents the voltage deviation from the desired level (e.g., 1.0\,p.u.).
    \item $\lvert q_{i}^{PV}\rvert$ penalizes the reactive power generated by PV inverters.
    \item $\alpha_{1}, \alpha_{2}$ are weighting factors balancing the trade-off between voltage stability and reactive power usage.
\end{itemize}

\textbf{Objective Function:} The objective of the reinforcement learning agent is to find a policy $\pi$ that maximizes the cumulative discounted reward:
\begin{equation}
    \label{eq:objective_appendix}
    \max_{\pi} \; \mathbb{E}_{\pi}
    \Bigl[\,\sum_{t=0}^{\infty}\gamma^{\,t}\,r_{t}\Bigr]
\end{equation}
where $\gamma$ is the discount factor, encouraging long-term stability and efficient voltage regulation.

This detailed setup ensures the environment accurately models real-world operational constraints and challenges in a PV-penetrated microgrid.

\subsection{Experiment Setup and Hyperparameters}

For the voltage regulation environment, \( \alpha_{1} = 1 \) and \( \alpha_{2} = 0.1 \). For the hyperparameters of both BCQ and CQL, please refer to Table~\ref{tab:hyperparams}. The experiment is implemented on a Windows system with an Intel 13th Gen i7 processor, an NVIDIA GeForce 4060ti GPU, and PyTorch 1.10.

\begin{table}[H]
    \centering
    \caption{Hyperparameters for BCQ and CQL}
    \label{tab:hyperparams}
    \begin{tabular}{lcc}
        \hline
        \textbf{Hyperparameter} & \textbf{BCQ} & \textbf{CQL} \\
        \hline
        Q-network learning rate & $10^{-4}$ & $10^{-4}$ \\
        Target network update rate ($\tau$) & 0.005 & 0.005 \\
        Batch size & 64 & 64 \\
        Discount factor ($\gamma$) & 0.99 & 0.99 \\
        Maximum action perturbation ($\Phi$) & 0.05 & -- \\
        Behavior cloning weight ($\lambda$) & 0.75 & -- \\
        VAE latent space dimension & 64 & -- \\
        Number of sampled actions per state & 10 & 10 \\
        Temperature parameter ($\alpha$) & -- & 0.1 \\
        Penalty weight for conservative loss ($\zeta$) & -- & 0.5 \\
        Regularization weight for policy constraint ($\beta$) & -- & 1.0 \\
        \hline
    \end{tabular}
\end{table}

\subsection{Dataset Statistics}

Table~\ref{tab:dataset_stats} lists several performance metrics for the Expert, Medium, and Poor datasets. These metrics offer insights into the three offline datasets’ variability, quality, and “experience level” (i.e., how much average valuable information each contains).

\vspace{-10pt} 
\begin{table}[H]

    \caption{Summary of Reward and Return Statistics for Expert, Medium, and Poor Datasets}
    \label{tab:dataset_stats}
    \centering
    \small 
    
    \begin{tabular}{lccc}
    \hline
     \textbf{Performance Metric} & \textbf{Expert} & \textbf{Medium} & \textbf{Poor} \\
    \hline
     Average Reward & -0.0242 & -0.0678 & -0.4325 \\
     Reward Variance & 1.97E-04 & 6.5995 & 7.6837 \\
     Reward Std. Deviation & 0.014 & 2.5691 & 2.227 \\
     Average Episode Return & -5.7464 & -15.8355 & -85.3963 \\
     Maximum Episode Return & -2.5771 & -13.75 & -20.349 \\
     Return Variance & 4.9976 & 1.44E+3 & 7.91E+3 \\
    \hline
    \end{tabular}
\end{table}

\subsection{Case Study on the IEEE 33-bus System}
\label{appendix:ieee33bus}

\begin{figure}[h]
    \centering
    \includegraphics[width=0.58\textwidth]{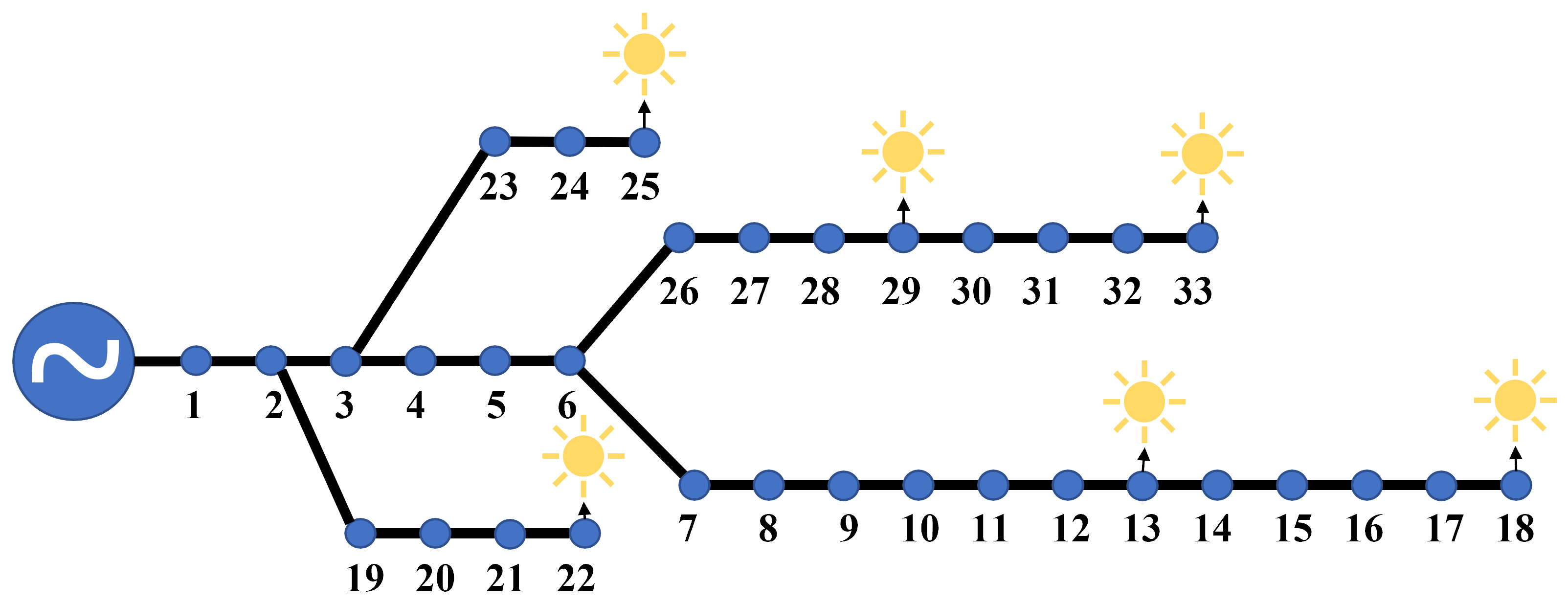}
    \caption{The IEEE 33-bus system with solar PV penetration.}
    \label{fig:ieee33bus_appendix}
\end{figure}

The IEEE 33-bus system is a widely used test case for distribution networks, representing a radial network structure with distributed loads and generation, which serves as the testbed for evaluating voltage regulation strategies in a PV-penetrated microgrid. Real-world load and PV generation data are used to simulate operational conditions, providing a comprehensive environment for assessing the performance of different offline-RL algorithms.

\textbf{Network Configuration:}
- Each load is represented by a blue circle in Fig.~\ref{fig:ieee33bus_appendix}.
- Specific nodes have installed photovoltaic (PV) generators, marked by solar symbols.

\textbf{Load Data:}
- The load data is derived from a 3-year dataset of Portuguese electricity consumption, capturing the usage patterns of 232 consumers.
- To introduce variability, the load profile is modified with a random perturbation of $\pm 5\%$ on the power factors.

\textbf{PV Generation Data:}
- The PV generation data is sourced from Elia, Belgium’s power network operator.
- It is adjusted to a 3-minute resolution to align with the control intervals, ensuring consistency with real-world operational conditions.

\textbf{Simulation Setup:}
This system provides a rigorous environment for evaluating different voltage regulation strategies under varying load and generation conditions. The realistic load data, PV penetration, and power flow simulations ensure the algorithms are tested in a representative operational setting.

\subsection{Performance on Offline Datasets}
\label{appendix:bcq_performance}

This section provides a detailed evaluation of BCQ's performance on the Expert, Medium, and Poor datasets, with corresponding metrics and visualizations.

\begin{figure}[h]
    \centering
    \includegraphics[width=0.5\textwidth, trim={0cm 0cm  0cm 0 cm}, clip]{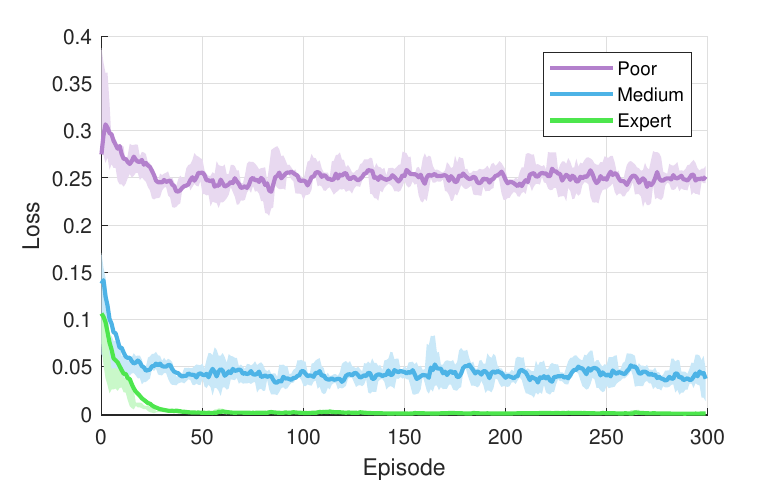}
    \caption{Comparison of the VAE total loss across datasets.}
    \label{fig:vae_loss_appendix}
\end{figure}

Fig.~\ref{fig:vae_loss_appendix} shows the VAE total loss for each dataset:
- Expert dataset: Exhibits a predictable and stable pattern, achieving the lowest reconstruction loss due to its high-quality transitions.
- Medium dataset: Has moderate reconstruction loss, reflecting its mixed-quality transitions with added noise.
- Poor dataset: Displays the highest reconstruction loss, indicating unstructured and variable data, consistent with its low-quality transitions.

\begin{figure}[h]
    \centering
    \includegraphics[width=1\textwidth, trim={3cm 0cm 3cm 0cm}, clip]{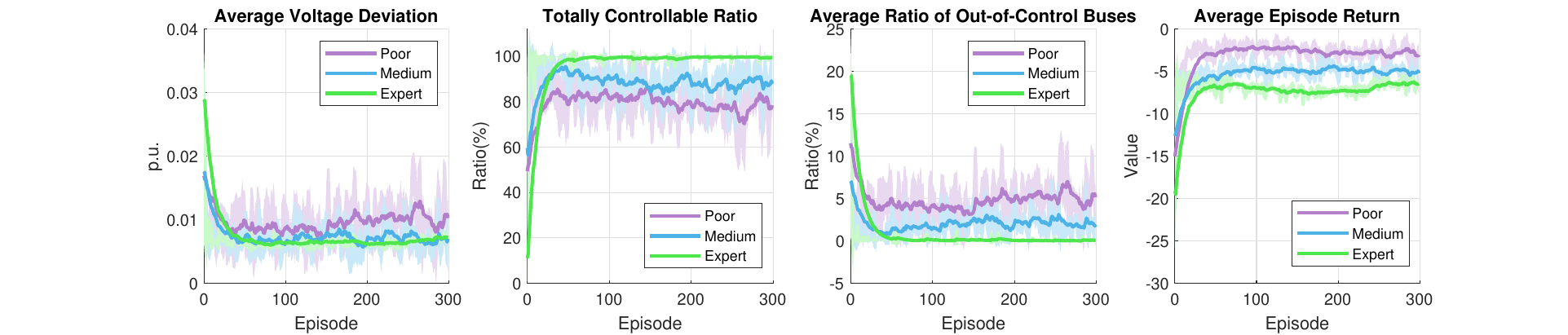}
    \vspace{-10pt}
    \caption{Voltage regulation performance across three offline datasets.}
    \label{fig:volt_reg_perf_appendix}
\end{figure}

Fig.~\ref{fig:volt_reg_perf_appendix} compares voltage regulation performance across the datasets using the following metrics:

1) \textbf{Totally Controllable Ratio}:  
   The percentage of steps where all buses maintain voltage levels within the desired range. Higher ratios indicate better controllability, with the Expert dataset achieving the highest ratios, followed by Medium and Poor datasets.

2) \textbf{Average Episode Return}:  
   Measures cumulative rewards per episode. Even with the Poor dataset, BCQ learns effective strategies and achieves reasonable returns.

3) \textbf{Average Voltage Deviation}:  
   Represents the deviation of bus voltages from the target level (1.0 p.u.). The Expert dataset yields the lowest deviation, followed by the Medium and Poor datasets.

4) \textbf{Average Ratio of Out-of-Control Buses}:  
   The proportion of buses outside the acceptable voltage range.
   The Expert dataset shows the smallest ratio, while the Poor dataset exhibits the highest ratio, demonstrating the influence of data quality.

These results highlight BCQ's ability to adapt to varying data quality levels while maintaining acceptable voltage regulation performance. However, its performance is directly influenced by the quality of the offline dataset, with lower-quality datasets leading to reduced controllability and stability.

\end{document}